\documentclass[10pt,twocolumn,letterpaper]{article}

\usepackage[pagenumbers]{cvpr} 

\usepackage{graphicx}
\usepackage{amsmath}
\usepackage{amssymb}
\usepackage{booktabs}

\usepackage[utf8]{inputenc} 
\usepackage[T1]{fontenc}    

\usepackage{url}            
\usepackage{booktabs}       
\usepackage{amsfonts}       
\usepackage{nicefrac}       
\usepackage{microtype}      
\usepackage{array}
\usepackage{floatrow}
\usepackage{multirow}
\usepackage{subcaption}
\usepackage{hhline}
\usepackage{wrapfig}
\usepackage{soul}
\usepackage[sort, numbers, compress, square]{natbib}
\usepackage[table,xcdraw]{xcolor}
\usepackage{siunitx}

\usepackage{algpseudocode} 
\usepackage{algorithm}
\usepackage{algcompatible}
\usepackage[colorinlistoftodos]{todonotes}
\usepackage{amsmath}
\usepackage{hhline}
\usepackage{blindtext}
\usepackage{amssymb}
\usepackage{enumitem}
\usepackage{framed}
\usepackage[linewidth=.5pt]{mdframed}
\usepackage{caption}
\usepackage{subcaption}
\usepackage{bm}

\usepackage{amsmath,amsfonts,bm,bbm}
\usepackage{xcolor}

\newcommand{\relu}{f_{\text{ReLU}}}

\def\eqref#1{equation~\ref{#1}}

\def\1{\bm{1}}

\DeclareMathAlphabet{\mathsfit}{\encodingdefault}{\sfdefault}{m}{sl}
\SetMathAlphabet{\mathsfit}{bold}{\encodingdefault}{\sfdefault}{bx}{n}

\usepackage[pagebackref,breaklinks,colorlinks]{hyperref}

\usepackage[capitalize]{cleveref}
\crefname{section}{Sec.}{Secs.}
\Crefname{section}{Section}{Sections}
\Crefname{table}{Table}{Tables}
\crefname{table}{Tab.}{Tabs.}

\def\cvprPaperID{10032} 
\def\confName{CVPR}
\def\confYear{2023}

\begin{document}

\title{\textit{NeMo}: 3D \textit{Ne}ural \textit{Mo}tion Fields from Multiple Video Instances of the Same Action}

\author{Kuan-Chieh Wang$^\dagger$, \ Zhenzhen Weng, \ Maria Xenochristou, \ Jo\~{a}o Pedro Ara\'{u}jo, \ Jeffrey Gu, \\
C. Karen Liu, \ Serena Yeung  \\
Stanford University \\
{\tt\small $^\dagger$wangkua1@stanford.edu} \\
{\small \url{https://sites.google.com/view/nemo-neural-motion-field}}
}
\maketitle

\begin{abstract}
The task of reconstructing 3D human motion has wide-ranging applications. The gold standard Motion capture (MoCap) systems are accurate but inaccessible to the general public due to their cost, hardware and space constraints.  In contrast, monocular human mesh recovery (HMR) methods are much more accessible than MoCap as they take single-view videos as inputs. Replacing the multi-view MoCap systems with a monocular HMR method would break the current barriers to collecting accurate 3D motion thus making exciting applications like motion analysis and motion-driven animation accessible to the general public. However, performance of existing HMR methods degrade when the video contains challenging and dynamic motion that is not in existing MoCap datasets used for training. This reduces its appeal as dynamic motion is frequently the target in 3D motion recovery in the aforementioned applications. Our study aims to bridge the gap between monocular HMR and multi-view MoCap systems by leveraging information shared across multiple video instances of the same action.  We introduce the Neural Motion (NeMo) field. It is optimized to represent the underlying 3D motions across a set of videos of the same action. Empirically, we show that NeMo can recover 3D motion in sports using videos from the Penn Action dataset, where NeMo outperforms existing HMR methods in terms of 2D keypoint detection.  To further validate NeMo using 3D metrics, we collected a small MoCap dataset mimicking actions in Penn Action,and show that NeMo achieves better 3D reconstruction compared to various baselines.

\end{abstract}

\section{Introduction}
\label{sec:introduction}

\begin{figure}
    \centering
    \includegraphics[width=\columnwidth,page=20, trim=0 0 470 0, clip]{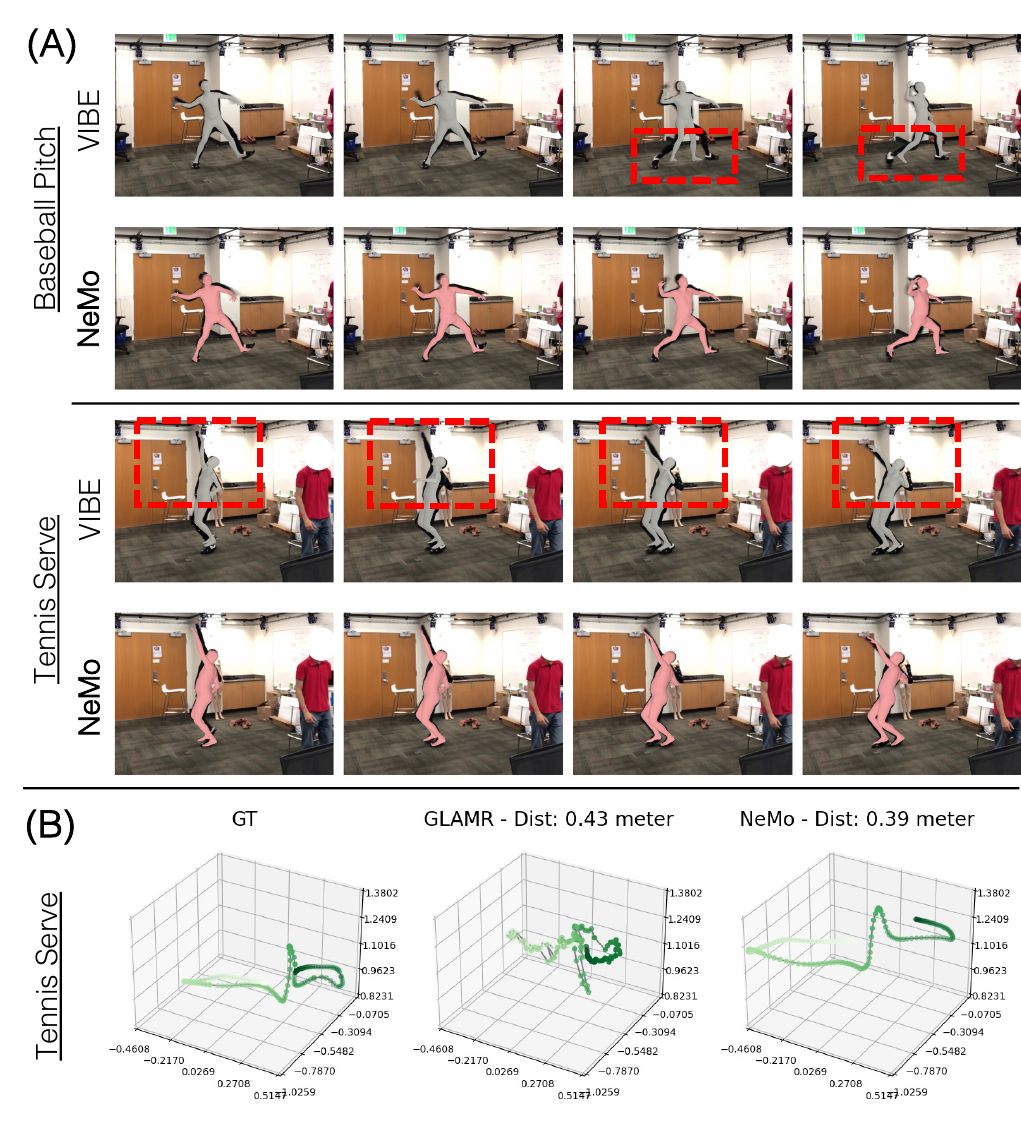}
    \caption{
    By leveraging the shared 3D information across videos, NeMo can accurately recover the \textit{dynamic} ranges of athletic motion, where existing 3D human mesh recovery (HMR) methods struggle.  To illustrate, VIBE, a baseline video-based HMR method, fails to capture the large step taken by the subject in the ``Baseball Pitch'' example, and swaps the arms in ``Tennis Serve'' (in subfigure A). Furthermore, NeMo recovers global root trajectory more accurately than existing global HMR method, GLAMR~\cite{yuan2022glamr} (in subfigure B).
    }
    \label{fig:fig1}
\end{figure}

Reconstruction of 3D human motion has wide-ranging applications from the production of animation movies like Avatar~\citep{cameron2009avatar}, realistic motion synthesis~\citep{xie2021physics,holden2016deep,xia2017survey} and biomechanical motion analysis~\cite{das2011quantitative,haralabidis2020fusing,uhlrich2022opencap,boswell2022smartphone}.
Existing MoCap systems are predominantly marker-based and work by recording 2D infrared images of light reflected by markers placed on the human subject.  
However, placing, calibrating and labelling the markers are all tedious processes, and the markers can potentially restrict the range-of-motion of the subject.  
There has been a recent development of ``markerless'' MoCap systems based on computer vision methods. 
For example, given a set of synchronized video captures from multiple views, one can run 2D keypoint detection methods like OpenPose~\citep{cao2019openpose} and perform triangulation to recover the 3D pose~\cite{nakano2020evaluation}.
Fundamentally, both MoCap approaches, marker-based and markerless, require \textit{multi-view} video capture of the same \textit{exact instance} of the motion.    

On the other hand, there has been a rapid development of 3D monocular human pose estimation (HPE) and human mesh recovery (HMR) methods. 
These methods aim to recover the 3D human motion from a single-view video capture.  
The accessibility of monocular HMR makes it an attractive alternative to MoCap systems.
However, monocular HMR is a challenging problem as a single-view input only provides partial information about the underlying 3D motion.  
The model needs to overcome complications like depth ambiguity and self-occlusion.  
HMR models, like other machine learning systems, overcome these difficulties by learning from paired training data.
However, paired video and MoCap datasets are scarce, and publically available MoCap datasets are often restricted to simple everyday motions like Human3.6M~\citep{ionescu2013human3} and AMASS~\citep{mahmood2019amass}.
As a result, existing HMR methods generalize less well in domains with less available MoCap data, such as motions in sports, a dominant application domain of 3D human motion recovery ~\cite{mundermann2006evolution,nagymate2018application,hamill2021biomechanics,colyer2018review}. See Figure~\ref{fig:fig1} for where existing HMR methods struggled to capture the dynamic range of athletic motions.

The multi-view assumption in MoCap is restrictive, but not having information from multiple views in HMR also makes the problem more challenging.  
We bridge the gap by assuming there is shared and complementary information in \textit{multiple instances} of video captures of the same action, similar to what is in the (same instance) multi-view setup.  
These \emph{multiple video instances} can be different repetitions of the same action from the same person, or even from different people executing the same action in different settings (See the left side of Figure~\ref{fig:nemo_method} for illustration).
A key feature of sports actions is that they are well-defined and structured.
For example, an athlete is also often instructed to practice by performing many repetitions of the same action and the variation across the repetitions is oftentimes slight.
Even when we look at the executions of the same action from different athletes, these different motion ``instances'' often contain shared information.
In this work, we aim to better reconstruct the underlying 3D motion from videos of multiple instances of the same action in sports. 

We parametrize each motion as a neural network. It takes as input a scalar phase value indicating the phase/progress of the action and an instance code vector for variation and outputs human joint angles, root orientation and translation. 
Since the different sequences are not synchronized, and the actions might progress at slightly different rates (e.g., a faster versus slower pitch), we use an additional learned phase network for synchronization.
The neural network is shared across all the instances while other components including the instance codes and phase networks are instance-specific.  All the components are learned jointly.
We optimize using the 2D reprojection loss with respect to the 2D joint keypoints of the input videos and prior 3D loss w.r.t. initial predictions from HMR methods to enforce 3D prior.  
We call the resulting neural network a Neural Motion (\textbf{NeMo}) field.
NeMo can also be seen as a new test-time optimization scheme for better domain adaptation of 3D HMR similar in spirit to  SMPLify~\citep{bogo2016keep,SMPL-X:2019}. A key difference is that we leverage shared 3D information at the group level of many instances to learn a canonical motion and their variations.  

To summarize, our contributions are:
\begin{itemize}
    \item We propose the neural motion (NeMo) field and an optimization framework that improves 3D HMR results by jointly reasoning about different video instances of the same action.  
    \item We optimize NeMo fields on sports actions selected from the Penn Action dataset~\citep{zhang2013actemes}. Since the Penn Action dataset only has 2D keypoint annotations, we collected a small MoCap dataset with 3D groundtruth where the actor was instructed to mimic these motions. 
    We show improved 3D motion reconstruction compared to various baseline HMR methods using both 3D metrics, and also improved results on the Penn Action dataset using 2D metrics.
    \item Our proposed NeMo field also recovers \emph{global} root translation. Compared to the recently proposed global HMR method, recovered global motion from NeMo is substantially more accurate on our MoCap dataset.
\end{itemize}

\section{Related Work}
\label{sec:related-work}
In this section, we discuss related work in 3D HMR methods, multi-view 3D modelling and human motion datasets.

\paragraph{HMR Methods}
Our proposed methode NeMo bridges the gap between monocular HMR methods~\citep{cho2022cross,zhang2022pymaf,guan2021bilevel,iqbal2021kama,sengupta2021hierarchical,kanazawa2018end,kanazawa2018learning,kanazawa2019learning}, and traditional multi-view MoCap systems.
In a way, it can be seen as a test-time optimization (TTO) extension for finetuning predictions from existing HMR methods, much like the popular TTO algorithm SMPLify~\cite{SMPL-X:2019}.  
Compared to SMPLify, NeMo leverages information across the multiple video instances of the same action, resulting in better 3D reconstruction.  
NeMo can be used in conjunction with any existing HMR methods that are video-based like  VIBE~\citep{kocabas2020vibe} or framed-based like PARE~\cite{Kocabas_PARE_2021}. 
Compared to most HMR methods, NeMo also recovers the global root trajectory, which is a central piece of MoCap data, while most HMR methods do not. 
Recently, \emph{global} HMR is attracting attention from researchers where global root trajectory is also recovered.  This is a more challenging but also more impactful version of the HMR task.  Compared to GLAMR~\cite{yuan2022glamr}, in terms of global HMR metrics, NeMo reduces the overall error in our experiments.  

\paragraph{Multi-view 3D Human Modeling}
Monocular HMR is fundamentally challenging due to issues such as occlusions and depth ambiguity. Multi-view 3D models aim to overcome these issues by utilizing video captures from multiple viewpoints to gain a holistic understanding of the scene. 
These multi-view videos could be shot changes of the same scene in movies~\citep{pavlakos2020human}, or different viewpoints recorded by multiple synchronized cameras~\citep{zhang2021direct,iqbal2020weakly,huang2021dynamic,7896626}.
In comparison, our approach uses different instances of the same action performed asynchronously by one or more humans to capture the 3D motion of a sports action. 
iMoCap~\citep{dong2020motion} studied a problem similar to ours by curating videos from the internet, and also aimed to recover the 3D motion.
In contrast to our neural representation, their method fitted a fixed set of poses over time, requiring them to additionally enforce temporal smoothness and cannot naturally allow for interpolations. Furthermore, their method does not leverage recent advances from monocular HMR, which is vital for having good 3D motion prior. 
Lastly, they curated their videos and did not use an existing video dataset, making comparison impossible.  In contrast, we apply NeMo to the Penn Action dataset and further validated it on a MoCap dataset we collected which we intend to open-source. 
\footnote{Only the raw videos were released for their project, but not the annotations, the extracted 3D motion or the code for their method.  Attempts to communicate with the authors were also unsuccessful.}

\paragraph{3D Human Motion Datasets}
Even though datasets with 3D ground truth human motion are essential to developing reliable models for human mesh reconstruction (HMR), collecting 3D data is costly and labor-intensive. The result is that available 3D datasets are limited in the number of subjects and motions. The Human3.6M dataset~\citep{ionescu2013human3} contains data (3D MoCap, 2D keypoints, action labels, and videos) for only 11 human subjects, while the 3DPW dataset~\citep{von2018recovering} is similar in having paired video and 3D groundtruth, but was captured in an outdoor environment, using a combination of inertial measurement units (IMUs) and vision models. Both the Human3.6M and 3DPW datasets do not contain sports motion. 
The AMASS dataset~\citep{mahmood2019amass} is much larger in scale, containing over 300 subjects and more than 11000 motions, but still inherits the restrictions of MoCap, and does not cover the full range of athletic motions. The lack of MoCap data for sports makes existing HMR methods suffer from the domain shift issue and perform poorly on sports sequences, especially during the dynamic segment of the motion.
Many sports datasets only contain 2D joint annotations and are significantly downsampled in time~\citep{zhang2013actemes,chen2021sportscap,andriluka2018posetrack}.

\section{Neural Motion (NeMo) Fields}
\label{sec:creating_3d}

\begin{figure*}
    \centering
    \includegraphics[width=\columnwidth,page=14, trim=0 260 0 0, clip]{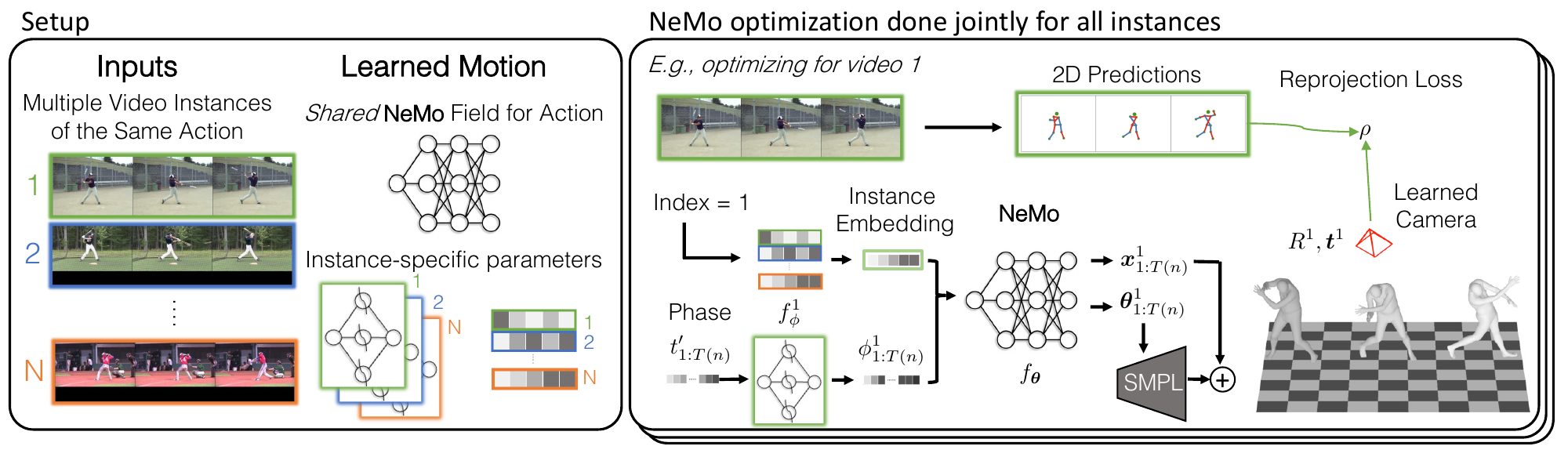}
    \caption{We propose a method for 3D global motion recovery that leverages shared information across multiple video instances of the same action, which we call NeMo.  Our method learns a shared canonical motion parametrized by a neural network, instance-specific phase networks and latent vectors.  Optimizing NeMo mainly relies on the 2D reprojection error. 
    The phase networks are monotonically increasing warping functions that help synchronize the different progressions across videos.  Given the warped phases along with a learnable instance embedding, the NeMo field outputs the joint angles and the root translation of the motion, which are rendered using the SMPL model~\citep{loper2015smpl}. 
    }
    \label{fig:nemo_method}
\end{figure*}

In this section, we focus on the problem of extracting the 3D human motion for specific athletic actions, such as ``baseball pitching'', given a set of videos.
We assume that for the same action, the underlying 3D human motion is similar across the videos. 
Intuitively, the 3D reconstruction task can be made easier by combining information from all the videos into a single motion with variations.
This makes the 3D reconstruction problem easier than treating all the videos separately.
See Figure~\ref{fig:nemo_method} for an illustration of our method.

\paragraph{Problem Formulation}
Given multiple video instances of the same action, our goal is to recover the 3D \emph{global motions}; namely, sequences of 3D poses (including the root orientation), $\bm{\theta}_{1:T}^n$, and root translations, $\bm{x}_{1:T}^n$ for each of the video instances.
The superscript $\circ^{n}$ denotes the $n$-th video instance and the subscript $\circ_{1:T}$ denotes a sequence from time $1$ to $T$.
Our key insight is that, for many actions, the variations across multiple instances (i.e, executions) can be slight, which means we can improve our estimate of the 3D motion by solving for the motion instances jointly.

We first process the videos using off the shelf 2D and 3D pose estimators to get the initial estimates of the 2D keypoints $\tilde{\bm{j}}_{1:T}^n$ and 3D poses $\tilde{\bm{\theta}}_{1:T}^n$.  We use tilde, $\tilde{\circ}$, to denote the initial estimates. 
We then try to optimize for the motion jointly using both the 2D and 3D initial predictions across all video instances of the same action.
Our method can also be viewed as a test-time optimization algorithm for improving 3D motion like SMPLify~\cite{SMPL-X:2019,kocabas2020vibe} that leverages the shared information at the group level.  
Note, most existing 3D HMR methods only output the pose/articulation (i.e., $\bm{\theta}$) and not the global root translation, $\bm{x}$.  
In contrast, we also aim to recover the 3D global root translation. 
In the following sections, we describe how we parametrize and optimize for the set of motions using a shared neural motion field.

\paragraph{Neural Motion Field} 
We represent a 3D motion sequence using a multi-layer perceptron (MLP) and call this representation a Neural Motion (NeMo) field.
The input to the network is the phase of the motion sequence, $\phi \in [0,1]$, which can be viewed as the current progression in a time series and an instance vector $\bm{z} \in \mathbb{R}^{N_z}$ to account for the instance variation of the motion. 
The instance vectors are also learnable parameters that are optimized jointly with NeMo.
The MLP outputs 23 joint angles, the root orientation $\bm{\theta}$ and 3D global translation $\bm{x}$, $f_{\bm{\Theta}}: \mathbb{R}^{1 + N_z} \mapsto \mathbb{R}^{24\times6 + 3}$.
The joints use the 6D representation for rotation proposed in Zhou et al. \cite{zhou2019continuity} makes optimizing angles easier, and is commonly used in HMR networks~\citep{kolotouros2019learning,kocabas2020vibe}.
For convenience, we denote the sub-network that outputs global translation as $f_{\bm{x}}$ and the rest that outputs joint angles and orientation as $f_{\bm{\theta}}$.

Given the output of NeMo fields, the joint angles and root translation, we use the Skinned Multi-Person Linear (SMPL) model~\citep{loper2015smpl} to represent the 3D mesh of the human body. The SMPL body model is a differentiable function $f_m: \mathbb{R}^{72+10} \mapsto \mathbb{R}^{6890\times3}$ that takes a pose parameter $\bm{\theta} \in \mathbb{R}^{72}$ and shape parameter $\bm{\beta} \in \mathbb{R}^{10}$, and returns the body mesh ${\bm{m}}$ with $6890$ vertices. 
In this work, we assume a neutral shape, which is fixed to the constant vector of zeros, i.e. $\bm{\beta} = \bm{0}$ and drops it in what follows for simplicity.
A linear regressor $W$ can be fitted to get the major body joints in 3D, $\bm{p} \in \mathbb{R}^{J\times 3}$ and  $\bm{p} = W \bm{m}$, where each joint is a linear combination of the mesh vertices. 
To get the 3D body joints given an input phase $\phi$, the combination of the NeMo field and SMPL is used as follows:
\begin{equation}
     \bm{p} = W\Big(f_{\bm{m}}\big(f_{\bm{\theta}}([\phi; \bm{z}])\big) +  f_{\bm{x}}([\phi; \bm{z}])\Big),
\end{equation}
where $[\cdot;\cdot]$ denotes concatenation.

\paragraph{Phase Networks} 
Since the videos are not synchronized, and the different motion instances can progress at different rates, we allow the phases for the different sequences to vary. We introduce a self-normalized monotonic neural network, $f_{\bm{\phi}}: \mathbb{R} \mapsto \mathbb{R}$, which takes as input the linearly normalized time index, $t' = \frac{t}{T}$, where $T$ is the total length of a given motion sequence and outputs the phase, $\phi$.  A monotonic neural network can be composed by summing $K$ shifted and scaled sigmoid function.  
The full phase network is written as:
\begin{align}
    \phi &= f_{\bm{\phi}}(t') = \frac{g(t') - g(0)}{g(1) - g(0)}, \label{eqn:self-norm}
\end{align}
where
\begin{align}
    g(t') &=\frac{1}{K } \sum_{k=1}^K \sigma( \relu(a_k)(t' - \relu(b_k))).
\end{align}
We define $\sigma(\cdot)$ as the logistic function, $\relu(\cdot)$ as the ReLU activation function, and  $\{a_k, b_k\}_{k=1}^{K}$ as the learnable shift and scale parameters.  
To ensure the phase starts at 0 and ends at 1, self-normalization is added (Equation~\ref{eqn:self-norm}).  
The ReLU function ensures the sigmoid functions are increasing.

\paragraph{NeMo Optimization}
NeMo optimization goes through two main stages.  In both stages, optimization is done jointly across all videos.  In the first stage, the pose component of the NeMo field (i.e., $f_{\bm{\theta}}$ is optimized w.r.t the initial 3D estimate $\tilde{\bm{\theta}}$ to mimic the prediction for the 3D pose estimator).  In the second stage, the warmed-up NeMo field, along with all the other parameters, are jointly optimized using 2D reprojection loss, which we describe below. 
In addition to a NeMo field, instance vectors and phase networks, we also fit the cameras.
Each camera has its own extrinsic parameters including a rotation matrix, $R$, a translation vector, $\bm{t}$, and intrinsic parameters.  We fix the intrinsic parameters and learn the extrinsic parameters; namely, how the cameras are placed in the 3D world. 
The optimization of NeMo can be written as:
\begin{align}
    &\min_{f_{\bm{\theta}}, f_{\bm{x}}, \{R^n, \bm{t}^n, f_{\bm{\phi}^n}, \bm{z}^n \}_{n=1}^N} \;\;\;\sum_{n=1}^N \Big( \frac{1}{T(n)} \sum_{t=1}^{T(n)} \rho(\bm{j}_t^n, \tilde{\bm{j}}_t^n) \Big),
\end{align}    
where,    
\begin{align}
    &\bm{j}_t^n = P\Big(R^n f_{3d}\big(\bm{p}^n_t , \bm{z}^n \big) - \bm{t}^n\Big), \\
    &\bm{p}^n_t = W\Big(f_{\bm{m}}\big(f_{\bm{\theta}}([\phi^n_t; \bm{z}^n])\big) +  f_{\bm{x}}([\phi^n_t; \bm{z}^n])\Big), \\
    & \phi^n_t = f^n_{{\bm{\phi}}}(\frac{t}{T(n)}).
\end{align}
We use $P$ to denote the perspective projection and $\rho(\cdot)$ the error function for 2D points. 
We use the Geman-McClure error function, which is more robust to outliers than the mean squared errors. $T(n)$ indicates the length of the $n$-th video.

\section{Experiments}
\label{sec:experiments}
\begin{table*}[t]
\centering 
\begin{tabular}{l|ccccc|c}
\toprule
Method       & \multicolumn{1}{c}{\begin{tabular}[c]{@{}c@{}}Baseball \\ Pitch\end{tabular}} & \multicolumn{1}{c}{\begin{tabular}[c]{@{}c@{}}Baseball \\ Swing\end{tabular}} & \multicolumn{1}{c}{\begin{tabular}[c]{@{}c@{}}Tennis \\ Serve\end{tabular}} & \multicolumn{1}{c}{\begin{tabular}[c]{@{}c@{}}Tennis \\ Swing\end{tabular}} & \multicolumn{1}{c}{\begin{tabular}[c]{@{}c@{}}Golf \\ Swing\end{tabular}} & \multicolumn{1}{|c}{Mean} \\
\midrule
& \multicolumn{5}{c}{\textit{MPJPE} (mm, $\downarrow$)}  &          \\
VIBE~\citep{kocabas2020vibe}         & 101.2 / 141.3          & 84.5 / 120.7          & 94.4 / 129.9          & 69.5 / 96.5           & 87.5 / 114.1          & 87.4 / 120.5          \\
VIBE+SMPLify~\citep{kocabas2020vibe} & 111.3 / 153.8          & 89.9 / 128.7          & 99.5 / 137.7          & 79.4 / 108.7          & 103.5 / 132.7         & 96.7 / 132.3          \\
GLAMR~\citep{yuan2022glamr}        & 99.7 / 131.8           & 100.4 / 139.3         & 116.0 / 155.1         & 80.2 / 106.3          & 114.0 / 150.2         & 102.1 / 136.5         \\
PARE~\citep{Kocabas_PARE_2021}         & 97.7 / 134.8           & 84.5 / 129.9          & 89.5 / 118.0          & 73.2 / 97.6           & 96.7 / 132.4          & 88.3 / 122.5          \\
NeMo (Ours)  & \textbf{85.8 / 108.7}  & \textbf{65.3 / 91.3}  & \textbf{80.6 / 97.9}  & \textbf{65.4 / 85.4}  & \textbf{78.9 / 94.2}  & \textbf{75.2 / 95.5}  \\
& \multicolumn{5}{c}{\textit{MPVPE} (mm, $\downarrow$)}  &          \\
VIBE~\citep{kocabas2020vibe}         & 126.7 / 178.8          & 101.1 / 149.2         & 117.6 / 164.4         & 86.5 / 122.4          & 108.9 / 141.7         & 108.2 / 151.3         \\
VIBE+SMPLify~\citep{kocabas2020vibe} & 139.5 / 196.2          & 108.8 / 160.1         & 124.8 / 174.5         & 101.5 / 141.7         & 123.9 / 157.6         & 119.7 / 166.0         \\
GLAMR~\citep{yuan2022glamr}        & 129.0 / 168.6          & 126.9 / 182.1         & 149.7 / 201.6         & 107.5 / 142.8         & 156.0 / 201.6         & 133.8 / 179.3         \\
PARE~\citep{Kocabas_PARE_2021}         & 122.8 / 170.4          & 102.5 / 163.9         & 113.5 / 150.8         & 93.6 / 127.8          & 121.2 / 163.6         & 110.7 / 155.3         \\
NeMo (Ours)  & \textbf{112.5 / 147.3} & \textbf{77.9 / 118.9} & \textbf{95.5 / 121.1} & \textbf{83.3 / 116.8} & \textbf{96.4 / 118.1} & \textbf{93.1 / 124.4}  \\
\bottomrule
\end{tabular}
\caption{\textbf{3D evaluation on our MoCap dataset.} Both errors over the entire sequence (\textit{left} in a cell) and over the dynamic range of a sequence (\textit{right} in a cell) are reported.  The improvement is more pronounced during the dynamic range of the motions where performance of existing HMR methods degrade.}
\label{table:eval_3d_merged}
\end{table*}
\begin{table*}[h]
\begin{tabular}{cr|
S[table-format=3.2]
S[table-format=3.2]
S[table-format=3.2]
S[table-format=3.2]
S[table-format=3.2]|
S[table-format=3.2]}
\toprule
                \textit{MoCap}             &       Method       & \multicolumn{1}{c}{\begin{tabular}[c]{@{}c@{}}Baseball \\ Pitch\end{tabular}} & \multicolumn{1}{c}{\begin{tabular}[c]{@{}c@{}}Baseball \\ Swing\end{tabular}} & \multicolumn{1}{c}{\begin{tabular}[c]{@{}c@{}}Tennis \\ Serve\end{tabular}} & \multicolumn{1}{c}{\begin{tabular}[c]{@{}c@{}}Tennis \\ Swing\end{tabular}} & \multicolumn{1}{c}{\begin{tabular}[c]{@{}c@{}}Golf \\ Swing\end{tabular}} & \multicolumn{1}{|c}{Mean} \\
\midrule
\multirow{2}{*}{Global-MPJPE (mm, $\downarrow$)} & GLAMR~\cite{yuan2022glamr}   & \textbf{144.01}         & 114.43         & 198.71       & \textbf{121.87}       & 128.26     & 141.46 \\
 & NeMo (Ours)   & 151.45         & \textbf{91.92}          & \textbf{146.06}       & 148.49       & \textbf{95.46}      & \textbf{126.68} \\
 \midrule
\multirow{2}{*}{Global-MPVPE (mm, $\downarrow$)} & GLAMR~\cite{yuan2022glamr}  & \textbf{159.78}         & 130.42         & 218.59       & \textbf{134.6}        & 158.13     & 160.3  \\
 & NeMo (Ours)  & 163.56         & \textbf{96.22}          & \textbf{149.36}       & 147.82       & \textbf{100.79}     & \textbf{131.55} \\
\bottomrule
\end{tabular}
\caption{\textbf{Corrected global 3D evaluation on our MoCap dataset.}}
\label{table:new_eval_3d_global}
\end{table*}
\begin{figure*}[!t]
    \centering
    \includegraphics[width=\columnwidth,page=12, trim=0 400 0 0, clip]{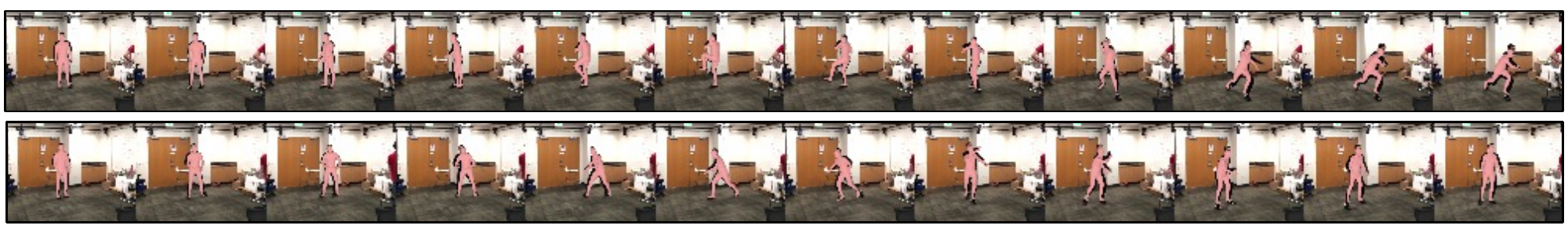}
    \caption{\textbf{Example rollout of our MoCap dataset.} Top is ``Baseball Pitch'' and bottom is ``Tennis Swing''. The rendered motions are from our learned NeMo fields. For rendered videos, please see the website.
    }
    \label{fig:mocap-rollout}
\end{figure*}
In this section, we validate our proposed method NeMo on two datasets: a MoCap dataset we collected, and the Penn Action dataset~\cite{zhang2013actemes}.  
We report standard 3D evaluation metrics for HMR and metrics for global HMR on our MoCap dataset (Section~\ref{sec:exp:mocap}). On Penn Action dataset where only 2D groundtruth is available, we report 2D metrics and show qualitative results (Section~\ref{sec:exp:penn}). Since our work focuses on dynamic and athletic motion, results are best viewed in videos. Please visit our supplemental project page for rendered results: \url{https://sites.google.com/view/nemo-neural-motion-field}. 

\paragraph{Methods} 
We used VIBE~\citep{kocabas2020vibe} for our initial 3D estimate and OpenPose~\cite{cao2019openpose} for our 2D psuedo-groundtruth. 
NeMo used an architecture of 3 hidden layer MLP.  We ran 300 steps using the 3D loss as warmup and 2000 steps using the 2D loss in the second stage.  The same hyperparameters were used for all motions across all datasets.  For more details please refer to Appendix~\ref{sec:app:experimental}. For baselines, we compared to the following HMR methods:
\begin{itemize}[parsep=2pt,itemsep=0pt] 
    \item \textbf{VIBE}~\citep{kocabas2020vibe} -- a video-based HMR method which we also used as our initial 3D estimate.
    \item \textbf{VIBE+SMPLify}~\citep{kocabas2020vibe,SMPL-X:2019} -- this method combines VIBE with SMPLify which finetunes the results using 2D reprojection loss.
    \item \textbf{PARE}~\citep{Kocabas_PARE_2021} -- a framed-based HMR method that trained on multiple datasets, including using 3D pseudo-groundtruth extracted by EFT~\citep{joo2020exemplar}. 
    \item \textbf{GLAMR}~\citep{yuan2022glamr} -- a state-of-the-art global HMR method that infer global root trajectory based on initial estimates from HybrIK~\cite{li2021hybrik}.
\end{itemize}

\paragraph{Penn Action Dataset} The Penn Action Dataset~\citep{zhang2013actemes} contains thousands of video sequences of different athletic actions with both action and 2D joint annotations for each sequence. 
We use this dataset as an example where 1. 3D groundtruth was not collected, and 2. using traditional MoCap to collect 3D groundtruth would be too expensive or infeasible because of constraints from the environment, motion, and availability of human experts. 

Specifically, we focus on five actions: ``Baseball Swing'', ``Baseball Pitch'', ``Tennis Serve'', ``Tennis Forehand'', and ``Golf Swing''.  The reason being that these actions are representative of our targeted problem: actions that are well defined and repeatable.  Other actions like ``Playing Guitar'' in the Penn Action dataset are not as well defined.

\paragraph{Our MoCap Dataset}
Since the Penn Action dataset only contains 2D keypoint groundtruth, and not 3D groundtruth, it is not enough for us to validate our reconstructed 3D motion.  
To validate our proposed method, we collected a MoCap dataset with their corresponding videos.  
We collected 8 repetitions/motion instances for each of the 5 actions above from different camera views.  The human actor is instructed to mimic the motion shown in the Penn Action dataset. See Figure~\ref{fig:mocap-rollout} for a visualization of the data.

\paragraph{Metrics}
The following metrics were used for evaluation.
\begin{itemize}[parsep=2pt,itemsep=0pt] 
    \item \textbf{MPJPE / MPVPE} -- mean per joint/vertex position error are commonly used for evaluating 3D HMR methods. MPJPE computes the distance from a predicted joint to the groundtruth joint in 3D and MPVPE computes distances for all vertices. Results are reported in millimeters (mm). 
    \item \textbf{Global-MPJPE / MPVPE} -- the global version of MPJPE/MPVPE measures the error taking into account the predicted global root translation and orientation.  This is in contrast to the non-global version where the prediction is root-centered.
    \item \textbf{2D Recon. Err.} -- 2D reconstruction error measures the 2D error predicted predicted and groundtruth joints in 2D.  Results are reported in terms of number of pixels.
    \item \textbf{PCK} -- percentage of correct keypoints is a measure of accuracy.  It threholds 2D reconstruction error by 10\% of the bounding box size of the target human.
\end{itemize}
More experimental details can be found in Appendix~\ref{sec:app:experimental}.

\subsection{Results on our MoCap Dataset}
\label{sec:exp:mocap}

In this section, we quantitatively evaluate the ability of NeMo to recover global 3D motion on our MoCap dataset. 
We compare NeMo to video-based HMR method with and without test-time optimization, state-of-the-art frame-based HMR method, and a recent global HMR method.  

\paragraph{Evaluation with Standard HMR Metrics} Table~\ref{table:eval_3d_merged} shows that NeMo outperforms baseline methods in terms of 3D metrics (MPJPE/MPVPE) across all actions.  The improvement is even more pronounced during the dynamic ranges of the motion~\footnote{See Appendix~\ref{sec:app:experimental} for the definition of dynamic range.}. 
This validates our original hypothesis that existing HMR methods are less robust for videos containing dynamic and athletic motion, which is an important application domain for 3D human motion recovery.
In the dynamic ranges of the motion, NeMo improve MPJPE from the best performing baseline VIBE from 120.5 mm to 95.5 mm, a 20.8\% improvement.  
Also worthnoting is the comparison with VIBE+SMPLify which also performs test-time optimization using 2D reprojection loss.
In a way, it can be seen as an ablation of NeMo that does not learn from multiple instances jointly.  
Interesting, while VIBE+SMPLify does not always improve the results from VIBE since the 2D keypoint predicted from OpenPose from a single video might not add more information.  
This stresses the importance of using the joint optimization proposed for NeMo.

In Appendix~\ref{sec:app:additional-results}, we include results using 2D evaluation metrics for the same experiment.
Note, while NeMo still performed the best overall, the performance between NeMo and baselines were much closer than they were in 3D evaluation because 2D projection is a lossy process and many erroneous 3D poses can be projected to the same 2D pose.
This speaks to the importance of using 3D evaluation for 3D motion recovery and our collected MoCap dataset.

\begin{figure}[!t]
    \centering
    \includegraphics[width=\columnwidth,page=18, trim=0 250 500 0, clip]{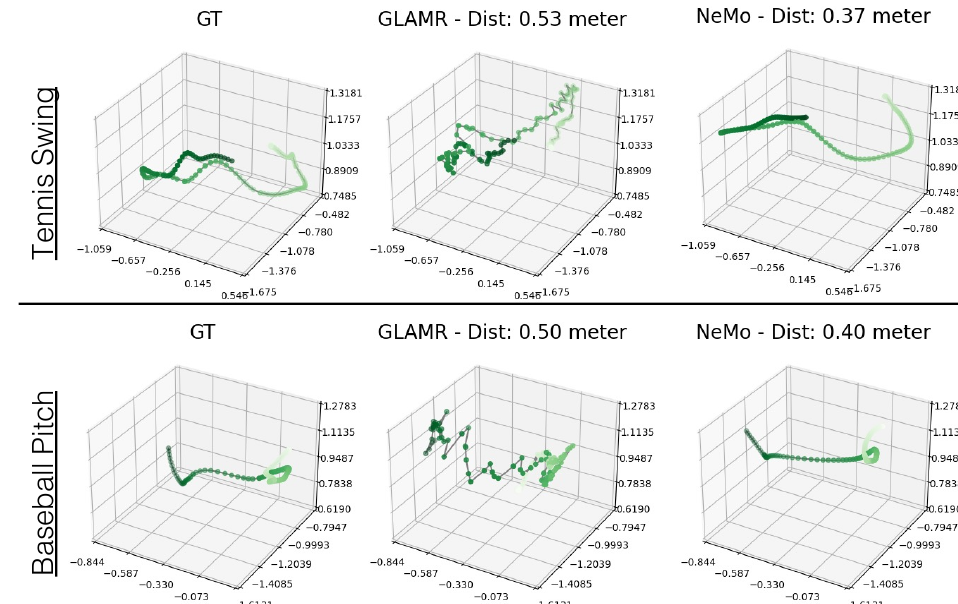}
    \caption{\textbf{Visualization of 3D global root trajectory on our MoCap dataset.} The brightness of the color denotes temporal progression (from dark to bright).}
    \label{fig:root}
\end{figure}

\paragraph{Evaluation with \emph{Global} HMR Metrics}
An important additional benefit of using NeMo is that the recovered motion contains global root information.  This is essential for applications like animation, viewpoint-free synthesis, and motion analysis. 
Global HMR is a recent task, and most of the baselines do not perform global HMR.  Comparing to GLAMR, a recent method for global HMR~\cite{yuan2022glamr}, NeMo improves the global MPJPE from 141 mm to  127 mm (see Table~\ref{table:new_eval_3d_global}). 
In Figure~\ref{fig:root} and~\ref{fig:fig1}B, we show example comparisons between recovered global root trajectory from NeMo and GLAMR. The results from NeMo is much less jittery and better represents the motion.  For example, in Figure~\ref{fig:fig1}B, the trajectory for the tennis serve captures the jump in the serving motion (i.e., the large increase and decrease in the z-axis). 
In Figure~\ref{fig:root}, we can see smooth and large steps taken during both of the actions from the NeMo motion, but not from the recovered motion using GLAMR.
Being able to capture the global movement is critical in the eventual goal of replacing MoCap systems with HMR methods.
Additional qualitative comparisons using rendered videos between NeMo and baselines are included in the website.

\begin{table*}[!t]\begin{tabular}{cr|
S[table-format=3.2]
S[table-format=3.2]
S[table-format=3.2]
S[table-format=3.2]
S[table-format=3.2]|
S[table-format=3.2]}
\toprule
              \textit{Penn Action}               &       Method       & \multicolumn{1}{c}{\begin{tabular}[c]{@{}c@{}}Baseball \\ Pitch\end{tabular}} & \multicolumn{1}{c}{\begin{tabular}[c]{@{}c@{}}Baseball \\ Swing\end{tabular}} & \multicolumn{1}{c}{\begin{tabular}[c]{@{}c@{}}Tennis \\ Serve\end{tabular}} & \multicolumn{1}{c}{\begin{tabular}[c]{@{}c@{}}Tennis \\ Swing\end{tabular}} & \multicolumn{1}{c}{\begin{tabular}[c]{@{}c@{}}Golf \\ Swing\end{tabular}} & \multicolumn{1}{|c}{Mean} \\

\midrule
\multirow{4}{*}{Recon. Err. ($\downarrow$)} & OpenPose~\cite{cao2019openpose}     & \textbf{12.71}          & 12.93          & 11.64        & 10.19        & 8.98       & 11.29 \\
                                            & VIBE~\cite{kocabas2020vibe}         & 14.2           & \textbf{7.35}           & 20.74        & 8.65         & \textbf{5.04}       & 11.2  \\
                                            & VIBE+SMPLify~\cite{kocabas2020vibe} & 16.15          & 20.56          & 9.74         & 20.15        & 7.23       & 14.77 \\
                                            & NeMo (Ours)  & 13.31          & 9.85           & \textbf{4.99}         & \textbf{4.36}         & 7.97       & \textbf{8.1}   \\
\midrule
\multirow{4}{*}{PCK ($\uparrow$)}           & OpenPose~\cite{cao2019openpose}     & \textbf{88.25}          & 88.91          & 92.1         & 93.93        & 93.81      & 91.4  \\
                                            & VIBE~\cite{kocabas2020vibe}         & 83.42          & 89.63          & 80.92        & 89.4         & 94.69      & 87.61 \\
                                            & VIBE+SMPLify~\cite{kocabas2020vibe} & 79.67          & 82.23          & 88.58        & 88.86        & 93.13      & 86.5  \\
                                            & NeMo (Ours)  & 80.16          & \textbf{90.93}          & \textbf{95.99}        & \textbf{97.46}        & \textbf{95.37}      & \textbf{91.98} \\
\bottomrule
\end{tabular}
\caption{\textbf{2D evaluation on the Penn Action dataset.}}
\label{table:penn_action_nemo_as_hmr}
\end{table*}
\subsection{Results on Penn Action}
\label{sec:exp:penn}

\begin{figure}[!t]
    \centering
    \includegraphics[width=\columnwidth,page=19, trim=0 100 490 0, clip]{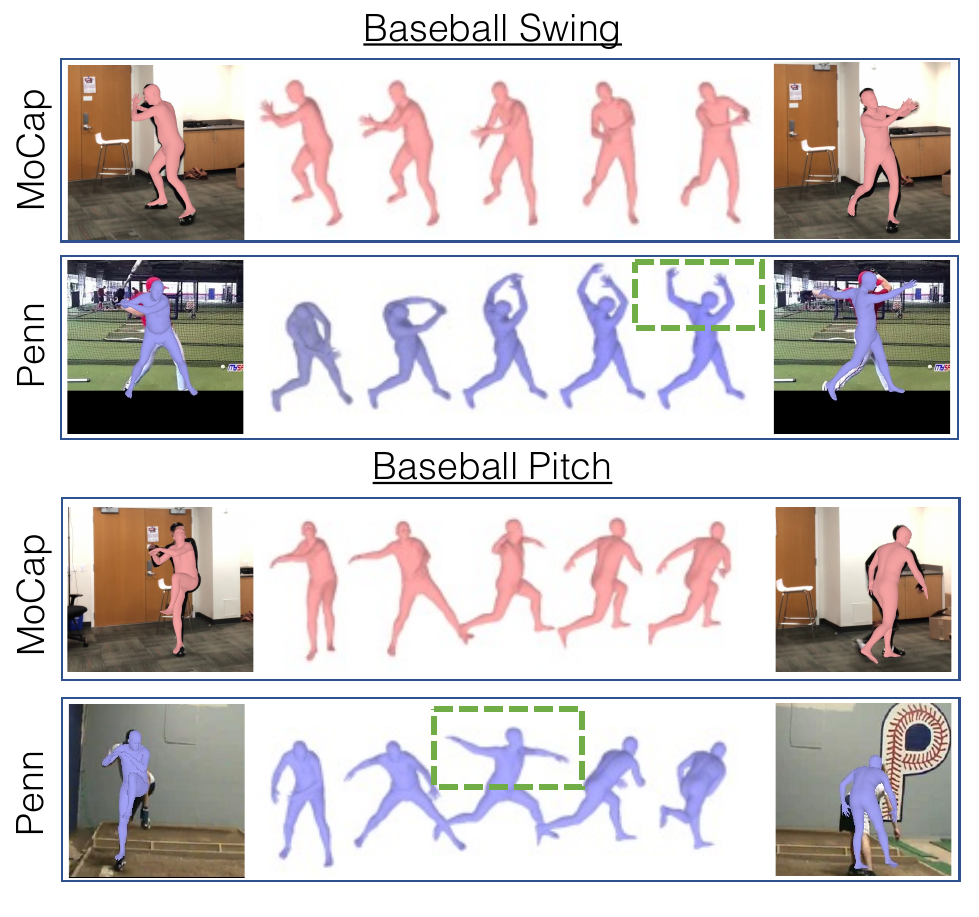}
    \caption{\textbf{Comparison of learned NeMo fields from our MoCap dataset and Penn Action.} Action in Penn Action is often executed by an advanced athlete whose motion is dynamic and exaggerated. Differences are highlighted by the green box.}
    \label{fig:datasets-side-by-side}
\end{figure}

\begin{figure}[!t]
    \centering
    \includegraphics[width=.8\columnwidth,page=15, trim=0 60 580 0, clip]{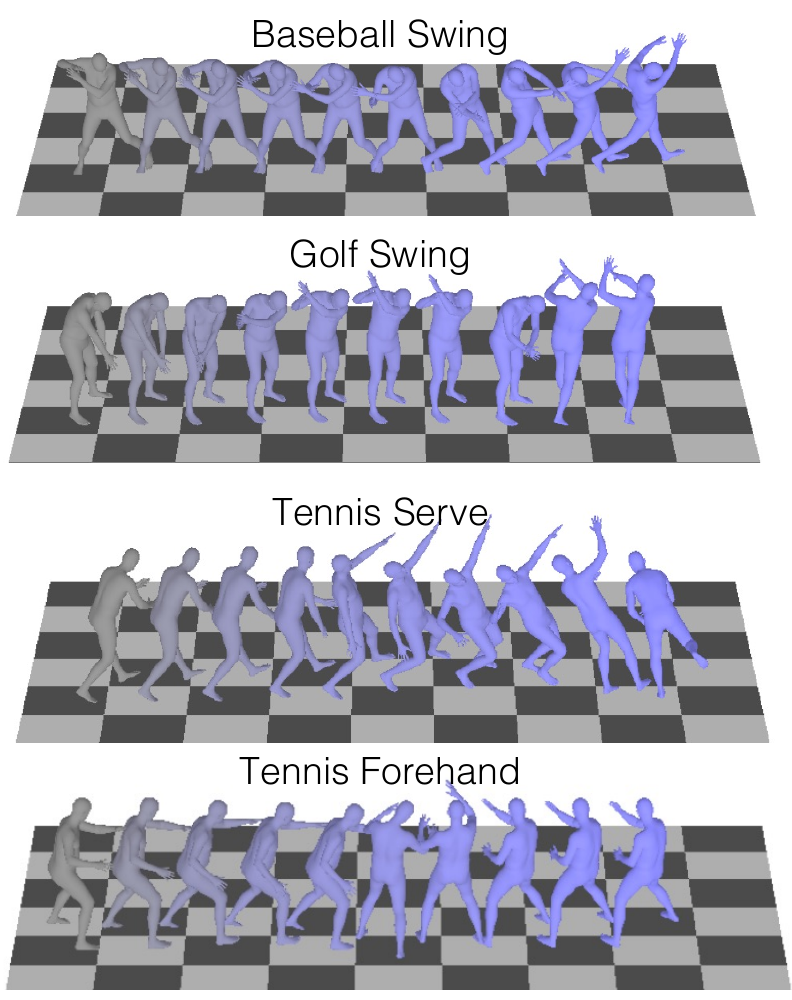}
    \caption{\textbf{Qualitative results on Penn Action.} Motion rollout from optimized NeMo fields on 4 actions from Penn Action. Actions progress from left to right.}
    \label{fig:actions}
\end{figure}

\paragraph{2D Evaluation}
The Penn Action dataset only has 2D keypoint annotations but not 3D groundtruth MoCap.  This is commonly the case as 2D annotations can be done post-hoc for most videos, but 3D MoCap can only be captured in a laboratory.  
We use the Penn Action dataset as a demonstration that NeMo can be applied to existing real world video captures.  
While we cannot validate the results using 3D metrics, Table~\ref{table:penn_action_nemo_as_hmr} shows that in terms of 2D metrics, the NeMo outperformed existing 2D and 3D pose estimators overall. 
To examine the realism of the recovered motion in 3D, we show qualitative results in the next paragraph.

\paragraph{Qualitative Results}
Figure~\ref{fig:actions} contains four instances of optimized NeMo fields on four distinct sports actions.  
Qualitatively, the optimized NeMo fields conform with our expectations of their respective actions and capture many details in their execution. 
Combined with the better performance in terms of 2D metrics in the previous paragraph, these results demonstrate the NeMo can be applied to real world videos like those in the Penn Action dataset and recover realistic 3D motion.
In Figure~\ref{fig:datasets-side-by-side}, we visualize the learned NeMo fields from Penn Action and our MoCap dataset beside each other.  One can qualitatively observe the difference in recovered motion.  Penn Action videos often capture advanced athletes whose motions are more dynamic and exaggerated compared to an amateur. Such as in the Baseball Swing motion, many players in Penn Action let go of their hands at the end of the swing whereas in our MoCap dataset the subject human did not.  In the Baseball Pitch motion, the player from Penn Action keeps their throwing arm back while they step forward, which gives their pitch more power.  
These results highlight the importance of being able to achieve motion recovery from in-the-wild videos.
Additional results that show more variations of the learned motion in 3D can be found in the Appendix~\ref{sec:app:additional-results}.

\begin{table}[t]
\begin{tabular}{l|ccc}
\toprule
                                                 
                    & \begin{tabular}[c]{@{}c@{}} MPJPE \\ ($\downarrow$) \end{tabular} & \begin{tabular}[c]{@{}c@{}} MPJPE \\ (Dynamic, $\downarrow$)\end{tabular} & \begin{tabular}[c]{@{}c@{}} 2D PCK \\ ($\uparrow$) \end{tabular} \\
\midrule
NeMo (full)               & 77.96    & 95.7                                                         & 99.08  \\
- 3D HMR            & 125.22   & 161.82                                                       & 98.4   \\
- Instance-specific & 134.85   & 194.6                                                        & 89.57 \\
\bottomrule
\end{tabular}
\caption{\textbf{Ablations.} The second row removes 3D supervision from the HMR initial estimates.  The last row removes the instance-specific learnable parameters. 
}
\label{table:ablations}
\end{table}

\paragraph{Ablations}
Table~\ref{table:ablations} shows that removing 3D loss using the initial predictions from the existing 3D HMR method hurts 3D reconstruction but the 2D metric can still appear good.  This shows the importance of using an initial 3D estimate to enforce a good 3D prior.  Removing the instance-specific parameters degrades the accuracy of the 3D reconstruction which is reflected in both the 3D and 2D metrics.

\section{Limitations \& Future Directions}
\label{sec:limitations}
One limitation of the NeMo model is the assumption of a fixed camera.  While there are many sport videos that have an almost still camera where NeMo is applicable, as shown in Section~\ref{sec:experiments}, many sports videos are captured with a moving camera. This is especially true for motion that covers a lot of grounds, like a volleyball spike, or a basketball layup.  Often the full action can only be captured by moving a camera to track the athlete. Extending NeMo to account for a moving camera will further improve its applicability.
Another worthy future direction is in using the learned NeMo fields as a data augmentation tools for improving regression-based HMR methods, similar to what was proposed in EFT~\cite{joo2021exemplar}.  Currently, NeMo works with multiple video instances of the same action and is a test-time optimization algorithm.  While it produces more accurate 3D results than existing HMR methods, it is slow and limited to repeatable actions.  Using it as a data collection tool to then finetune HMR methods can potentially lead to a more accurate HMR methods that is also efficient.

\section{Conclusion}
\label{sec:conclusion}
We proposed NeMo, a neural motion representation and an optimization framework for extracting 3D motions given a set of different videos instances of the same sports action. 
Compared to existing HMR methods whose performance degrade in sports videos due to domain shift, NeMo can better recover the 3D motion of athletic motion by leveraging shared information across different video instances.
To validate NeMo, we collected a MoCap dataset mimicking the Penn Action dataset and show that NeMo outperformed a range of HMR baselines -- frame-based, video-based, test-time optimization algorithm, and global HMR method. We also evaluated NeMo on the Penn Action dataset using 2D metrics, and show qualitative results.  
Furthermore, NeMo can recover a much more faithful 3D root trajecotry when compared to a recently proposed global HMR method.

This project falls under the umbrella of works that aim to improve 3D reconstruction in the wild using computer vision.
Collectively, our society has already built a massive database of videos that capture the human experience in the form of movies, sports event broadcasts, news media and more.  
Technology that can transform this existing data to their 3D reconstruction will take us closer to a realistic virtual experience.
By using existing videos, we might even reconstruct events in the past, like Michael Jordan winning his first NBA title in 1991, and see it happen from anywhere on the basketball court.

{\small
\bibliographystyle{ieee_fullname}
\bibliography{arxiv_main}
}

\newpage
\onecolumn
\appendix
\begin{center}
\Large
\bf
Appendix
\end{center}

\section{Experimental Details}
\label{sec:app:experimental}
\paragraph{Model Details }  We used VIBE~\citep{kocabas2020vibe} for our initial 3D estimate and OpenPose~\cite{cao2019openpose} for our 2D psuedo-groundtruth. 
NeMo used an architecture of 3 hidden layer MLP with 1000 hidden units. The phase networks consists of 100 sigmoids nodes.  The instance code dimension was 5. In addition to 2D reprojection loss, during the optimization a regularization w.r.t. the VPoser and GMM pose prior were also used similar to SMPLify~\citep{SMPL-X:2019}. We ran 300 steps using the 3D loss as warmup and 2000 steps using the 2D loss in the second stage. The optimizer used for our optimization was Adam with a learning rate of $0.0001$ using the default second order hyperparameters in PyTorch.

\paragraph{``Dynamic Range''}
Dynamic range is defined as the segment of motion where the maximum joint velocity in 3D is above 2 m/s.  A contiguous segment is selected based on the first and last frame of the video that satisfied this criterion.

\paragraph{Metrics} For the standard MPJPE and MPVPE, the evaluation is done in a root-centered fashion at the frame-level, meaning that the root translation and orientation were aligned with the groundtruth at every frame. For global MPJPE and MPVPE, since each prediction resides in a different frame-of-reference, results were first aligned using rigid-body transformation (i.e., translation and orientation) with the groundtruth using vertices of the entire sequences.  Unlike Procrustes alignment which is sometimes used for HMR studies, we did not perform ``scaling''.

\paragraph{Penn Action Dataset}
We annotated each action with a ``left-handed'' or ``right-handed'' label, and only put action with the same handedness in the same batch. 
In the following experiments, for each action we sampled 40 batches of 3 sequences randomly from the training set of each action label.  For each sequence, we uniformly sampled 50 frames from the beginning to the end of action.  
The 2D annotations in the Penn Action dataset is noisy.  The joints are sometimes mislabelled. 
To alleviate this issue, we run OpenPose~\cite{cao2019openpose} on the videos.  
If the OpenPose prediction of a joint and the groundtruth label are more further than a threshold, we drop that keypoint in our optimization.  The threshold is set to 10\% of the image dimension.

\section{Additional Results}
\label{sec:app:additional-results}
\begin{table*}[h]
\begin{tabular}{cr|
S[table-format=3.2]
S[table-format=3.2]
S[table-format=3.2]
S[table-format=3.2]
S[table-format=3.2]|
S[table-format=3.2]}
\toprule
                \textit{MoCap}             &       Method       & \multicolumn{1}{c}{\begin{tabular}[c]{@{}c@{}}Baseball \\ Pitch\end{tabular}} & \multicolumn{1}{c}{\begin{tabular}[c]{@{}c@{}}Baseball \\ Swing\end{tabular}} & \multicolumn{1}{c}{\begin{tabular}[c]{@{}c@{}}Tennis \\ Serve\end{tabular}} & \multicolumn{1}{c}{\begin{tabular}[c]{@{}c@{}}Tennis \\ Swing\end{tabular}} & \multicolumn{1}{c}{\begin{tabular}[c]{@{}c@{}}Golf \\ Swing\end{tabular}} & \multicolumn{1}{|c}{Mean} \\

\midrule
\multirow{5}{*}{Recon. Err. ($\downarrow$)} &    OpenPose~\cite{cao2019openpose}          & 32.74          & 25.7           & 50.5         & 25.46        & 37.54      & 34.39 \\
                             & VIBE~\cite{kocabas2020vibe}         & 18.17          & 15.28          & 20.08        & 13.88        & \textbf{14.5}       & 16.38 \\
                             & VIBE+SMPLify~\cite{kocabas2020vibe} & 18.09          & 15.46          & 26.61        & 13.89        & 15.04      & 17.82 \\
                             & PARE~\cite{Kocabas_PARE_2021}         & 16.06          & 15.19          & \textbf{16.26}        & \textbf{13.16}        & 15.45      & 15.23 \\
                             & NeMo (Ours)  & \textbf{15.7}           & \textbf{14.67}          & 16.48        & 13.81        & 15.12      & \textbf{15.16} \\
\midrule
\multirow{5}{*}{PCK ($\uparrow$)}         & OpenPose~\cite{cao2019openpose}     & 95.77          & 97.62          & 94.26        & 98.12        & 96.15      & 96.38 \\
                             & VIBE~\cite{kocabas2020vibe}         & 97.64          & 98.51          & 96.96        & 99.55        & \textbf{99.34}      & 98.4  \\
                             & VIBE+SMPLify~\cite{kocabas2020vibe} & 97.65          & 98.56          & 95.93        & 99.51        & 99.25      & 98.18 \\
                             & PARE~\cite{Kocabas_PARE_2021}         & \textbf{99.33}          & 98.92          & \textbf{99.23}        & 99.71        & 98.51      & 99.14 \\
                             & NeMo (Ours)  & 98.46          & \textbf{99.61}          & 98.81        & \textbf{99.88}        & 99.16      & \textbf{99.18} \\
\bottomrule
\end{tabular}
\caption{\textbf{2D evaluation of our MoCap dataset.}}
\label{table:eval_2d}
\end{table*}
Table~\ref{table:eval_2d} shows 2D evaluation on our MoCap dataset.  While NeMo still performed the best overall, the trend is not as clear as in using 3D metrics (Table~\ref{table:eval_3d_merged}).  This is because many incorrect 3D poses can be reprojected to the same 2D joint locations.  First, it is worth noting that NeMo improves in terms of 3D evaluation while still outperforming baselines in terms of 2D metrics overall.  Second, the discrepancy between the 2D and 3D evaluations speaks to the importance of using 3D evaluation for quantitative results, and also checking results visually.

\end{document}